\documentclass[letterpaper, 10 pt, conference]{ieeeconf}

\IEEEoverridecommandlockouts                            
                                                        
\usepackage{graphicx}
\usepackage{amsmath}
\usepackage{amssymb}
\usepackage{color}
\usepackage{subfigure}

\graphicspath{ {./img/} }

\renewcommand{\v}[1]{\mathbf{#1}}

\title{\LARGE \bf Temporal changes in stimulus perception improve
  bio-inspired source seeking}

\author{A. Peque\~no-Zurro \and D. Shaikh \and I. Ra\~n\'o% <-this % stops a space
  \thanks{A. Peque\~no-Zurro, D. Shaikh, and I. Ra\~n\'o are with the Embodied Systems for Robotics and %
    Learning Unit, M\ae rsk McKinney M\o ller Institute, University of Southern Denmark, Denmark. %
    e-mail: {\tt\small\{alz, danish, igra\}@mmmi.sdu.dk}}}

\begin{document}

\maketitle
\thispagestyle{empty}
\pagestyle{empty}

\begin{abstract}
  Braitenberg vehicles are well known qualitative models of sensor
  driven animal source seeking (biological taxes) that locally
  navigate a stimulus function.  These models ultimately depend on the
  perceived stimulus values, while there is biological evidence that
  animals also use the temporal changes in the stimulus as information
  source for taxis behaviour. The time evolution of the stimulus
  values depends on the agent's (animal or robot) velocity, while
  simultaneously the velocity is typically the variable to
  control. This circular dependency appears, for instance, when using
  optical flow to control the motion of a robot, and it is solved by
  fixing the forward speed while controlling only the steering
  rate. This paper presents a new mathematical model of a bio-inspired
  source seeking controller that includes the rate of change of the
  stimulus in the velocity control mechanism. The above mentioned
  circular dependency results in a closed-loop model represented by a
  set of differential-algebraic equations (DAEs), which can be
  converted to non-linear ordinary differential equations (ODEs) under
  some assumptions. Theoretical results of the model analysis show
  that including a term dependent on the temporal evolution of the
  stimulus improves the behaviour of the closed-loop system compared
  to simply using the stimulus values. We illustrate the theoretical
  results through a set of simulations.
\end{abstract}

\section{Introduction}

Braitenberg vehicles \cite{braitenberg84vehicles} are a well known
model of animal navigation in robotics. They constitute a set of
bio-inspired navigation primitives \cite{rano14biologically} to move
towards or away from stimulus, known in biology as taxes
\cite{fraenkel61orientation}. Whilst in the seminal work of
Braitenberg these models defined steering control principles based on
direct perception of the stimulus, there is evidence in the literature
showing that animals also use the temporal change of the perceived
stimulus to drive their movement \cite{hellwig16rising}. However, the
dynamics of the stimulus depends on the movement of the animal, as for
the same static stimulus faster speed implies faster perceived
temporal changes. In this paper we develop a new mathematical model of
Braitenberg vehicle 3a, that achieves target seeking, and includes
steering control based on the temporal evolution of the perceived
stimulus. Because these wheeled vehicles rely on the unicycle motion
model, the resulting closed-loop equations are a set of non-linear
differential-algebraic equations, which can be converted (under some
conditions) to standard non-linear differential equations. The formal
analysis of the model allows to identify the equilibrium points and
some stability properties of this special controller, and shows that
it generates better trajectories compared to the original Braitenberg
3a vehicle. To the best of our knowledge this work presents the first
attempt to analyse theoretically a velocity-based sensor driven
navigation mechanism that depends on the velocity of the robot
itself. This work has implications for navigation mechanism based on
information dependent on the temporal changes like the optical flow or
event cameras, where the perception depends on the movement of the
robot.

Braitenberg vehicles are typically used in the literature with
unconventional sensors (i.e. sensors other than range or positioning
sensors) to implement local navigation using, among others; sound
(phonotaxis), light (phototaxis) or chemical (chemotaxis) stimuli, but
there are also use cases for local navigation using range sensors.
The work in \cite{bernard10phonotaxis} presents a model of the rat's
peripheral and central auditory system that enables the authors to
implement phonotaxis through a Braitenberg vehicle 3a to control the
motion of a mobile robot.  A model of the auditory system of lizards
was presented in \cite{shaikh09braitenberg} \cite{shaikh16from} to
control a mobile robot performing phonotaxis using Braitenberg vehicle
2b.  The phonotaxis behaviour of female crickets towards male crickets
is mimicked in a series of papers \cite{webb01spiking}
\cite{horchler04robot} \cite{reeve05new} that use the principles of
Braitenberg vehicles 2a and 3b to design spiking neural
networks. Their experimental results show that their robot achieves
excellent performance even in outdoor scenarios with the use of whegs
(wheeled legs). One of the earlier works in robotic chemotaxis
(localisation of an odour source) analysed experimentally the
behaviour of vehicles 3a and 3b \cite{lilienthal04experimental}. While
the connection between the sensors and the wheels was linearly
proportional, the chemical concentration readings were continuously
normalised due to the saturation and dynamic effects of the chemical
sensing technology.

The recent development and use of other unconventional sensing
modalities has also lead to the use of Braitenberg vehicles and their
principles on non-wheeled robots. A fish robot endowed with pressure
sensors imitating the fish lateral line used Braitenberg vehicle 2b to
implement rheotaxis \cite{salumae12against}, the alignment of the fish
body with a current or flow. Compared to other control mechanisms, the
deviation of the robot using this bio-inspired mechanism with respect
to the flow direction was significantly lower. Moreover, because the
movement of the fish-robot affected the measurements of the lateral
line, this work constitutes an experimental example of the control
mechanism staying stable under movement induced sensing dynamics.
Another application of the principles of Braitenberg vehicles to fish
robotics is the work presented in \cite{lebastard12underwater}. In
this work, the motion of an electric fish-robot is controlled by the
differences between the currents sensed by electrodes located on the
sides of the robot. The robot moves towards conductive objects in a
pond, avoiding isolating obstacles. Similarly to the previous fish
robot, the movement occurs in two dimensions, yet in the case of the
electric fish the robot is rigid and controlled through a rod. Another
example where Braitenberg vehicles are applied to non-wheeled robots
is the work presented in \cite{rano2018bridging}, where a simulated
snake robot is controlled to approach a light source by generating
modulated undulatory signals in its joints.  One type of
unconventional sensor that has recently gained traction is the event
cameras, which are used in \cite{milde17obstacle} in combination with
a Braitenberg vehicle 2b to implement visual obstacle avoidance in a
wheeled robot using neuromorphic hardware.

As we mentioned, Braitenberg vehicles are often used with
unconventional sensors, although some instances can be found that use
range sensors instead. In \cite{bicho97dynamic} a Braitenberg vehicles
were used for target acquisition of a light source -- phototaxis
through vehicle 3a --, and infrared sensor based obstacle avoidance --
through vehicle 2b. This work shows that proximity sensors can be also
used in conjunction with this bio-inspired local navigation
technique. Furthermore, this work inspired an implementation of a
Braitenberg vehicle based on the readings of a laser scanner to
estimate the free area around a robot \cite{rano14bio}. This paper
also proves that the motion of vehicle 2b can lead to chaotic
trajectories, a feature exploited in \cite{rano17biologically} to
implement area coverage in a simulated environment with a stimulus
similar to the free area around a robot. The chaotic coverage strategy
outperforms random walks and Levy walks.

As we saw, Braitenberg vehicles have been used with multiple sensor
and different robot types (mainly wheeled robots) but always using the
originally proposed instantaneous sensor readings or low pass filtered
signals. The main contribution of this paper is to extend this
successful control mechanism with a controller dependent on the
dynamic changes of the stimulus perceived by the vehicle. Furthermore,
we analyse the stability of the resulting controller and show that it
is stable and outperforms the original model. This work is inspired by
the biological findings supporting the use of rates of change of
perception for control and the successful experimental results
obtained in simulations motivated by these biological findings
\cite{pequeno18bio}.  The rest of the paper is organised as follows.
Section~\ref{sec:Model} reviews the definitions and standard
assumptions of Braitenberg vehicle 3a and presents the new control
mechanism that includes the temporal change of the stimulus. This
section also analyses the stability of the model under some
assumptions. Section \ref{sec:simulations} presents a series of
simulations that illustrate the main theoretical results
obtained. Finally, section \ref{sec:conclusions} presents some
conclusions and future work.

\section{Dynamic model of Braitenberg vehicle 3a}
\label{sec:Model}

\begin{figure}[h!]
  \begin{center}
    \includegraphics[width=0.7\columnwidth]{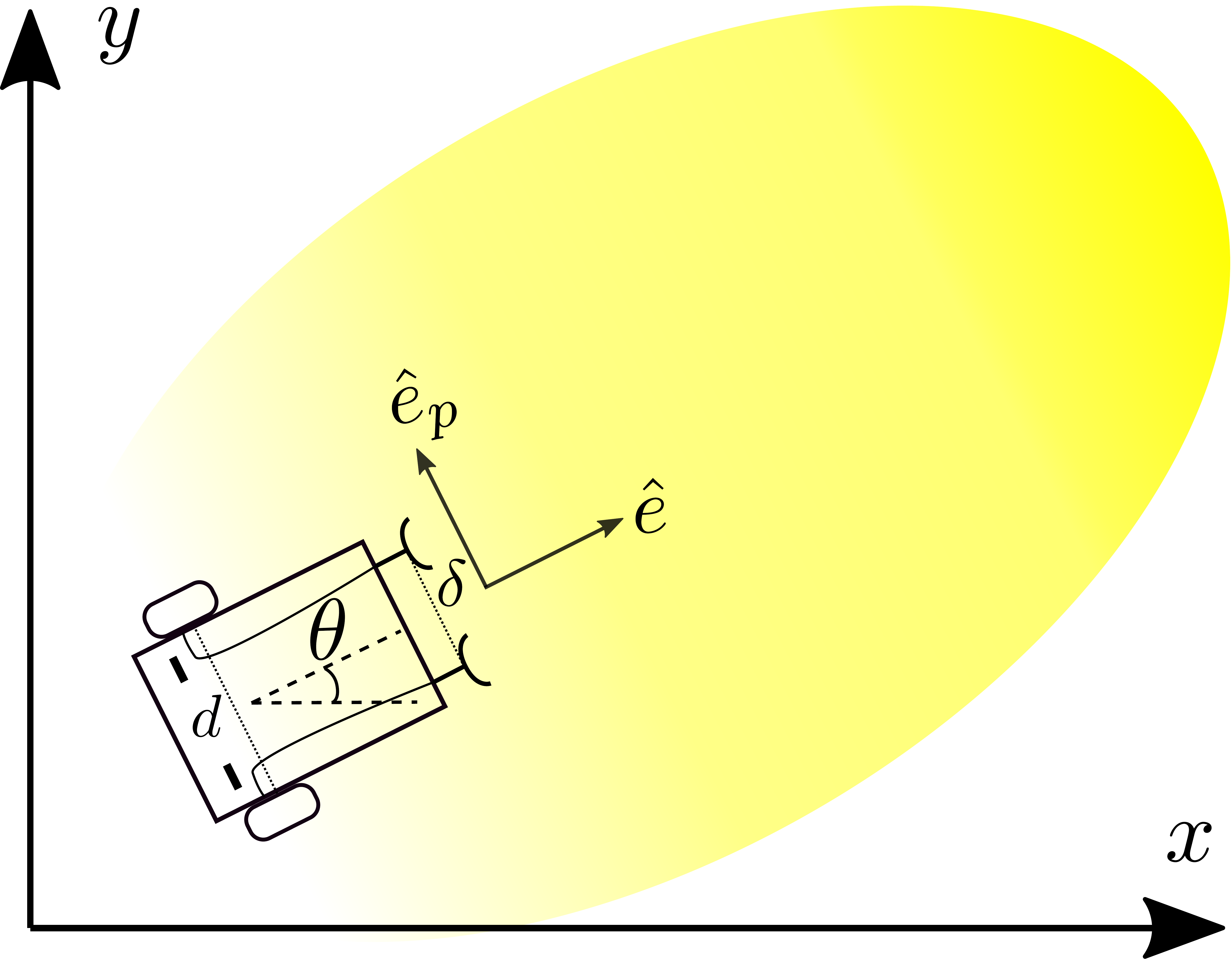}
    \caption{Representation of the qualitative model of Braitenberg
      vehicle 3a}
    \label{fig:BV}
  \end{center}
\end{figure}

Let's assume a Braitenberg vehicle 3a with two sensors and a direct
connection between the sensors and the motors as shown in
figure~\ref{fig:BV}. The state of the vehicle in the 2D plane can be
represented by the vector $\v{X}=[x,y,\theta]$, and we will denote
just the Cartesian coordinates by $\v{x}=[x,y]$. The selected $x$ and
$y$ coordinates correspond to the middle point between the sensors,
and we define the orthogonal reference system linked to the front of
the vehicle as $\hat{e}=[\cos\theta,\sin\theta]$ and
$\hat{e}_p=[-\sin\theta,\cos\theta]$.

We assume a smooth positive stimulus $S(\v{x})$ exists in the plane,
i.e. $S(\v{x})>0$ for all $\v{x}\in\Re^2$ and $S(\v{x})$ is of class
$C^\infty$. Without loss of generality we can assume that the maximum
of the stimulus occurs at $\v{x}=\v{0}$, i.e. $S(\v{0})>S(\v{x})$ for
all $\v{x}\in\Re^2$.  Defining the distance between the sensors is
$\delta$, we can express the positions of the right and left sensors
as $\v{x}_r=\v{x}-\frac{\delta}{2}\hat{e}_p$
$\v{x}_l=\v{x}+\frac{\delta}{2}\hat{e}_p$ respectively. Therefore, the
value of $S(\v{x})$ measured by these sensors will be $S(\v{x}_r)$ and
$S(\v{x}_l$.  According to the standard model of vehicle 3a
\cite{rano09steering} the velocity of each wheel is a decreasing
function $F(s)$ of the stimulus perceived by the sensor located on the
same side of the vehicle. We will assume the vehicle only moves
forward $F(s)>0$, while $F(s)$ being decreasing implies $F'(s)<0$.
Therefore, the velocity of each wheel is in general a non-linear
function of the stimulus $v=F(S(\v{x}))$ (or $v=F(\v{x})$ for
short). Because we want the vehicle to reach the maximum and stop
there, $F(s)$ has to be chosen such that $F(S(\v{0}))=0$.  However, in
this paper we will consider the case in which the velocity also
depends on the temporal rate of change of the stimulus,
i.e. $v=F(s,\dot{s})$, and as a first step we will assume that the
contribution of the stimulus derivative is additive, which allows is
to state the velocities of each wheel as:
\begin{eqnarray}
  v_r&=&F(S(\v{x}_r))+G(\dot{S}(\v{x}_l)) \nonumber\\
  v_l&=&F(S(\v{x}_l))+G(\dot{S}(\v{x}_r))
  \label{eq:wheelsV}
\end{eqnarray}
where $G(\dot{s})$ can be a non-linear function with the constraint
$G(0)=0$ for the vehicle to stop at the maximum.  It is worth noting
that the effects on the speed of the temporal derivative of the
stimulus is contralateral, i.e. derivative of $S(\v{x})$ on the right
sensor effects the velocity of the left wheel, and vice-versa.

Substituting the expressions for $\v{x}_r$ and $\v{x}_l$ in equations
(\ref{eq:wheelsV}), we can state the speeds of the right and left
wheels ($v_r$ and $v_l$) as a function of the state of the vehicle
(the Cartesian coordinates $\v{x}$ and its orientation $\theta$) and
the distance between the sensors $\delta$. Furthermore, we can
approximate the compound functions $F(S(\cdot))$ and $G(S(\cdot))$ as
first order Taylor series around $\v{x}$ assuming $\delta$ is small
relative to changes in $S(\v{x})$, and then compute the forward speed
of the vehicle, $v=\frac{v_r+v_l}{2}$, and its turning rate
$\omega=\frac{v_r-v_l}{d}$, where $d$ is the wheelbase. This leads to
the following speeds:
\begin{eqnarray}
v&=&F(S(\v{x}))+G(\nabla S^T\dot{\v{x}}) \nonumber\\
\omega&=&-\frac{\delta}{d}\nabla F(S(\v{x}))^T\hat{e}_p\nonumber\\
  &&\quad-\frac{\delta}{d}G'(\nabla S^T\dot{\v{x}})
  \left[\dot{\theta}\nabla S^T\hat{e}
    -\hat{e}_p^THS\dot{\v{x}}\right]
  \label{eq:VW}
\end{eqnarray}
where $\nabla F(S(\v{x}))$ is the gradient of the compound function
$F(S(\v{x}))$, $\nabla S$ and $HS$ denote the gradient and Hessian
matrix of the stimulus respectively, and we used the fact that
$\dot{S}(\v{x})=\nabla S^T\dot{\v{x}}$.

Introducing equations (\ref{eq:VW}) in the unicycle model we obtain
the closed-loop equations of motion of the dynamic Braitenberg vehicle
3a:
\begin{eqnarray}
\dot{x}&=&F(S(\v{x}))\cos\theta+G(\nabla S^T\dot{\v{x}})\cos\theta\nonumber\\
\dot{y}&=&F(S(\v{x}))\sin\theta+G(\nabla S^T\dot{\v{x}})\sin\theta\nonumber\\
\dot{\theta}&=&-\frac{\delta}{d}\nabla F(S(\v{x}))\hat{e}_p\nonumber\\
    &&-\frac{\delta}{d}G'(\nabla S^T\dot{\v{x}})
    \left[\dot{\theta}\nabla S^T\hat{e}
      -\hat{e}_pHS\dot{\v{x}}\right]
    \label{eq:bv3aDyn}
\end{eqnarray}

It is worth noting that while the model of Braitenberg vehicle 3a is
stated as a system of non-linear ordinary differential equations,
equations (\ref{eq:bv3aDyn}) correspond to a differential-algebraic
system of equations (DAE) as the time derivative of the state
$\dot{\v{X}}=[\dot{\v{x}},\dot{\theta}]$
($\dot{\v{x}}=[\dot{x},\dot{y}]$) is on the r.h.s of the equations.
DAEs are much more complex to solve than ODEs, as, for instance, the
set of meaningful initial conditions could be restricted. Therefore,
we will make some assumptions that will allow us to analyse the
closed-loop behaviour analytically.

\subsection{From Differential Algebraic to Ordinary Differential Equations}
Deriving analytic results from equations (\ref{eq:bv3aDyn}) can be
challenging, however, the DAEs can be converted into a system of
ordinary differential equations if we assume that the function
$G(\dot{S})$ is linear, i.e $G(\dot{S})=\alpha\dot{S}$, which is the
simplest function fulfilling the condition $G(0)=0$. It can be seen
that the closed-loop equations can be stated as
$\dot{\v{X}}=\mathcal{F}(\v{X})+\alpha A(\v{X})\dot{\v{X}}$, where the
derivative of the state on the r.h.s is multiplied by the matrix:
\begin{eqnarray}
  A&=&
  \left[
    \begin{array}{ccc}
      S_x\cos\theta & S_y\cos\theta &0 \\
      S_x\sin\theta & S_y\sin\theta &0 \\
      \frac{\delta}{d}\hat{e}_p^THS|_x&
      \frac{\delta}{d}\hat{e}_p^THS|_y&
      -\frac{\delta}{d}\nabla S^T\hat{e}
    \end{array}
    \right]\nonumber
\end{eqnarray}

where $S_x=\frac{\partial S}{\partial x}$, $S_y=\frac{\partial
  S}{\partial y}$, and $\hat{e}_p^THS|_x$ $\hat{e}_p^THS|_y$ represent
the $x$ and $y$ components of the row vector $\hat{e}_p^THS$, the
result of multiplying $\hat{e}_p^T$ by the Hessian matrix of
$S(\v{x})$. The vector flow $\mathcal{F}(\v{X})$ is formed from the
terms on the r.h.s. of the equation not involving the time derivatives
of the state $\v{X}$.

The motion equations of the vehicle (\ref{eq:bv3aDyn}) can be stated
as $\dot{\v{X}}=[I-\alpha A(\v{X})]^{-1}\mathcal{F}(\v{X})$, which is
an ordinary differential equation, assuming the matrix $I-\alpha A$ is
not singular. Therefore, the determinant of $I-\alpha A$:
\begin{eqnarray}
  |I-\alpha A|&=&-\frac{1}{d}\left[\alpha\nabla S^T\hat{e}-1\right]
  \left[\alpha\delta\nabla S^T\hat{e} + d\right]
\end{eqnarray}
should be different than zero, which provides us with the conditions
under which the DAE (\ref{eq:bv3aDyn}) can be turned into the ODE
$\dot{\v{X}}=[I-\alpha
  A(\v{X})]^{-1}\mathcal{F}(\v{X})$. Specifically, the determinant
vanishes when $\nabla S^T\hat{e}=\frac{1}{\alpha}$ or when $\nabla
S^T\hat{e}=-\frac{d}{\delta\alpha}$. Interestingly, if $\delta=d$ both
conditions can be stated as $|\nabla S^T\hat{e}|=\frac{1}{\alpha}$
(assuming $\alpha>0$, which will be justified below), i.e. the inverse
of $\alpha$ should be different from the directional gradient in the
direction of motion along the trajectory of the vehicle. This leads to
a constrain for the possible values of $\alpha$, namely
$\alpha\ne\frac{1}{\nabla S^T\hat{e}}$. If this condition is
fulfilled, the closed-loop motion equations corresponding to the
dynamic Braitenberg vehicle 3a are:
\begin{eqnarray}
\dot{x}&=&\frac{F(S(\v{x}))\cos\theta}{1-\alpha\nabla S^T\hat{e}}\nonumber\\
\dot{y}&=&\frac{F(S(\v{x}))\sin\theta}{1-\alpha\nabla S^T\hat{e}}\nonumber\\
\dot{\theta}&=&-\frac{\delta\nabla F^T\hat{e}_p}
    {(d+\alpha\delta\nabla S^T\hat{e})}\nonumber\\
    &&-
    \frac{\alpha\delta F(S(\v{x}))\hat{e}_p^THS\hat{e}}
               {(\alpha\nabla S^T\hat{e}-1)(d+\alpha\delta\nabla S^T\hat{e})}
               \label{eq:Model}
\end{eqnarray}

These equations are similar to the closed-loop model of Braitenberg
vehicle 3a \cite{rano09steering}, in fact the original model can be
obtained by simply setting $\alpha=0$. Moreover, it can be shown that
they have an equilibrium point at the origin -- the maximum of
$S(\v{x})$ -- since $\nabla S(\v{0})=\v{0}$ and $F(S(\v{0}))=0$, which
implies $\dot{x}=0$, $\dot{y}=0$ and $\dot{\theta}=0$.  The analysis
of the stability of the equilibrium point in the case of Braitenberg
vehicle 3a is presented elsewhere, but for the case at hand the
stability properties could change due to the additional term in the
equation for $\dot{\theta}$.

The condition on the determinant of $I-\alpha A(\v{X})$ for the
inverse matrix to exist is also reflected in the equations, as when
$|\nabla S|=-\frac{1}{\alpha}$ or $|\nabla S|=-\frac{d}{\alpha\delta}$
the denominators of the three dynamic equations would go to zero
generating possibly unbounded speeds. So the parameter $\alpha$ has to
be chosen carefully considering the possible values of the norm of the
stimulus gradient. While the equation for the angular variable
$\dot{\theta}$ is hard to analyse at first sight, the equations for
$\dot{x}$ and $\dot{y}$ provide a new insight on the effect of the
dynamic controller over the closed-loop behaviour. First it is worth
noting that the forward speed of the vehicle is
$v=\frac{F(S(\v{x}))}{1-\alpha\nabla S^T\hat{e}}$, where $\hat{e}$ is
the vector pointing in the direction of the vehicle and $\nabla S$
points in the direction of increasing stimulus. If the vehicle points
exactly in the direction of $\nabla S$, the dot product of the two
vectors corresponds to the norm of the gradient, i.e. $\nabla
S\hat{e}=|\nabla S|$, and if it points in the direction opposite to
the stimulus gradient it will be $\nabla S\hat{e}=-|\nabla S|$. Now
let's assume $\alpha>0$ and define $\nabla S^*$ as the gradient with
the largest norm in the environment or the domain where $S(\v{x})$ is
defined.  If we impose the additional condition on $\alpha$
$0<\alpha<\frac{1}{|\nabla S^*|}$ the term $\alpha\nabla S^T\hat{e}$
will be positive and smaller than one when the vehicle points in the
direction of increasing $S(\v{x})$. The overall effect compared to the
Braitenberg vehicle 3a is to increase the forward speed since the
denominator of $v$ becomes smaller than one. On the other hand, if the
vehicle points in the direction opposite to the gradient, the dot
product $\nabla S^T\hat{e}$ will be negative, making the denominator
of the forward speed larger than one, which effectively reduces the
forward speed. This is a highly interesting effect of introducing the
temporal rate of change of the stimulus in the velocity control of the
vehicle, as it makes the vehicle to move faster when it faces the
source, while reduces its speed when the source is on its
back. Analysing the behaviour of the the equation for $\dot{\theta}$
is not trivial since in the dynamic case there is a new term which
could change the sign of the original rotation speed (for which
stability conditions can be derived \cite{rano14results}). In the next
section we will take a closer look at the stability of the system
under the additional assumption of parabolically symmetric stimulus.

\subsection{Motion under parabolic stimulus}
\label{sec:parab}
In this section we will assume the stimulus has parabolic symmetry of
the form $S(\v{x})=S(\v{x}^TD\v{x})$ where $D=diag(\sigma_x,\sigma_y)$
is a positive definite diagonal matrix. This will allows us to analyse
the stability and derive more results on the trajectories of the
closed-loop controller.  The assumption holds for any smooth stimulus
close enough to the source. Moreover, since the reference system can
be defined arbitrarily we can rotate the $x$ and $y$ axis to be
aligned with the principal axis of the stimulus.  Under these
assumptions the gradient of the stimulus can be written as $\nabla
S=2S'(\v{x}^TD\v{x})D\v{x}$, where $S'(\cdot)$ is the derivative of
$S(\cdot)$ w.r.t its argument, with $S'(\cdot)<0$ because the stimulus
to grows towards the source (located at $\v{x}=\v{0}$). The Hessian
matrix of this stimulus is
$HS(\v{x})=4S''(\v{x}^TD\v{x})D\v{x}\v{x}^TD+2S'(\v{x}^TD\v{x})D$
which is negative definite at the origin, as the origin is a
maximum. Using the additional condition on the control function
$F(S(\v{0}))=0$ and evaluating $\dot{x}$, $\dot{y}$ and
$\dot{\theta}$, eq. (\ref{eq:Model}) at the stimulus maximum we obtain
the following condition for an equilibrium point to appear
$\hat{e}_p^TD\hat{e}=0$, which is satisfied for all $\theta$ if
$\sigma_x=\sigma_y$, or for $\theta=0$, $\theta=\pi$ and
$\theta=\pm\pi/2$ if $\sigma_x\ne\sigma_y$. So the closed-loop system
has infinite equilibrium points for circularly symmetric stimuli
($\sigma_x=\sigma_y$), or four equilibrium points for parabolically
symmetric stimuli $\sigma_x\ne\sigma_y$.

From the ODE form of the closed-loop equations we can also find an
analytic solution to the equations. Just like the standard Braitenberg
vehicle, when the initial pose of the vehicle falls on, and is aligned
with, the main axis of the stimulus (lines $x=0$ and $y=0$ with
heading $\theta=0$ or $\theta=\pi$ and $\theta=\pm\frac{\pi}{2}$
respectively) the vehicle follows a straight line trajectory towards
(or away from) the maximum of the stimulus. Taking, for instance
$x(0)=-x_0$, $y(0)=0$ and $\theta(0)=0$ as a starting pose we get for
the $y$ coordinate that $\dot{y}=0$ and for the $x$ coordinate:
\begin{eqnarray}
  \dot{x}&=&\frac{F(S(\sigma_xx^2))}{1-2\alpha S'(\sigma_xx^2)\sigma_xx}
  \label{eq:dotX}
\end{eqnarray}
while the equation of $\theta$ becomes $\dot{\theta}=0$ since $\nabla
F$ is perpendicular to $\hat{e}_p$ and $\hat{e}_pHS\hat{e}=0$ (for
$\theta=0$ $\hat{e}=[1,0]$ and $\hat{e}_p=[0,1]$). Therefore, the
trajectory is indeed a straight line with the evolution of $x$ given by
equation (\ref{eq:dotX}).

Having this particular solution to the system of non-linear equations
(\ref{eq:Model}) allows us to linerarise the system around that
trajectory and to check the evolution of the nearby trajectories with
increments $\Delta\v{X}$ following
$\Delta\v{\dot{X}}=M(t)\Delta\v{X}$. The linearised system is a time
varying linear system, and the eigenvalues of the matrix $M(t)$ --
which depend on the solution of equation (\ref{eq:dotX}) -- provide
information on the stability of the non-linear system, i.e. first
Lyapunov criterion applied to a non-constant trajectory. Because the
linearised system is three dimensional, at least one eigenvalue will
be real, while the other two could be complex conjugates, which is the
case when $\alpha=0$, i.e. the standard Braitenberg vehicle under
parabolically symmetric stimuli \cite{rano14results}. If we denote
$F=F(S(\sigma_x x^2))$, $F'=F'(S(\sigma_x x^2))$, $S'=S'(\sigma_x
x^2)$, $S''=S''(\sigma_x x^2)$, $\Delta_1=1-2\alpha\sigma_xxS'$ and
$\Delta_2=d-2\alpha\delta\sigma_xxS'$ the real eigenvalue of the
matrix $M(t)$ can be written as:
\begin{eqnarray}
  \lambda_1&=&\frac{2\sigma_xxF'S'}{\Delta_1}
  +\frac{2\alpha\sigma_xF[S'+2\sigma_xx^2S'']}{\Delta_1^2} \nonumber 
\end{eqnarray}
while the real part of the complex conjugate eigenvalues is:
\begin{eqnarray}
  Re[\lambda_{2,3}]&=& \frac{\delta\sigma_xxF'S'}{\Delta_2}
  + \frac{\alpha\delta F[S'[\sigma_x-\sigma_y]+2S''\sigma_x^2x^2]}
  {\Delta_1\Delta_2} \nonumber 
\end{eqnarray}
where $\sigma_x>0$, $x<0$, $F>0$, $F'<0$ and $S'<0$. It can be seen
that a sufficient condition for the real the eigenvalue to be negative
is $S'+2\sigma_xx^2S''<0$, but $\lambda_1$ could be negative even if
this condition is not met. Moreover, because $0<\Delta_1<1$ from the
condition imposed on $\alpha$, $\lambda_1$ will be more negative than
the corresponding eigenvalue for the standard Braitenberg vehicle,
which implies a faster convergence rate towards the straight line
solution. While the first term of the real part of the eigenvalues
$\lambda_{2/3}$ can be seen is negative, a similar analysis leads to
the necessary condition
$S'(\sigma_x-\sigma_y)+2S''\sigma_x^2x^2<0$. As $x$ approaches zero
from the negative starting point in the linear solution of equation
(\ref{eq:dotX}), the condition becomes $S'(\sigma_x-\sigma_y)<0$.

\section{Simulations}
\label{sec:simulations}
This section presents simulations to illustrate the theoretical
results on the proposed controller mechanism presented in the previous
section. Specifically we first show the accuracy of the model, to then
illustrate the improved stability when including a component with the
time derivative of the stimulus. It is worth noting that simulating
the differential-algebraic equations requires special integration
algorithms and consistency checks of the initial
conditions. Specifically, the algorithm used in the simulations below
is the one presented in \cite{shampine02solving}. All the results
assume a parabolically shaped stimulus with a maximum at the origin
and linear control function $F(s)$ to generate an equilibrium point at
$\v{x}=\v{0}$.

\subsection{Model evaluation}
\begin{figure}[h!]
  \begin{center}
    \includegraphics[width=\columnwidth]{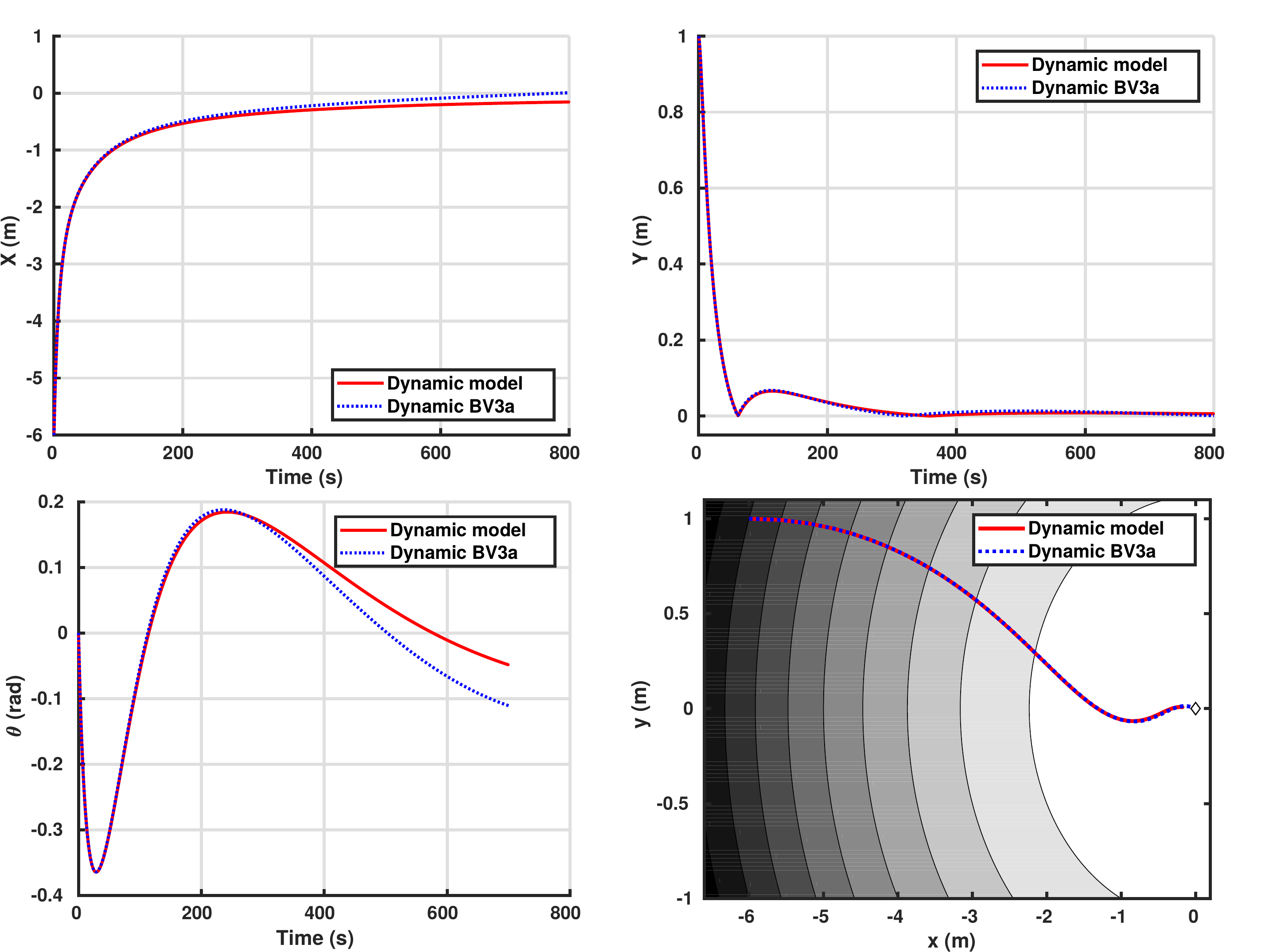}
    \caption{Simulation to evaluate the model accuracy with starting
      poses $x=-6$, $y=1$ and $\theta=0$} \label{fig:1}
  \end{center}
\end{figure}

To evaluate the accuracy of the model, eqs. (\ref{eq:bv3aDyn}), of the
dynamic Braitenberg vehicle -- which uses the values on the middle
points between the sensors --, we simulated it along with 
equations (\ref{eq:wheelsV}) to compute the trajectory. Because
equations (\ref{eq:bv3aDyn}) are obtained after first order truncation
of the Taylor series for $F(S(\v{x}))$ and $G(\dot{S}(\v{x}))$ in
equations (\ref{eq:wheelsV}), we can expect results of both
simulations to differ slightly.  Figure~\ref{fig:1} shows the
evolution over time of the $x$, $y$ (top row) and $\theta$
coordinates, together with the trajectory on the $x-y$ plane (bottom)
for $\delta=d=0.25$. The initial pose for both simulations was $x=-6$,
$y=1$, and $\theta=0$, and as the figures show than trajectories are
very similar. It is worth noting that the approximation error
increases with the distance between the sensors ($\delta=0.25$).

\subsection{Forward velocity increase towards the source}
\begin{figure}[h!]
  \begin{center}
    \includegraphics[width=0.9\columnwidth]{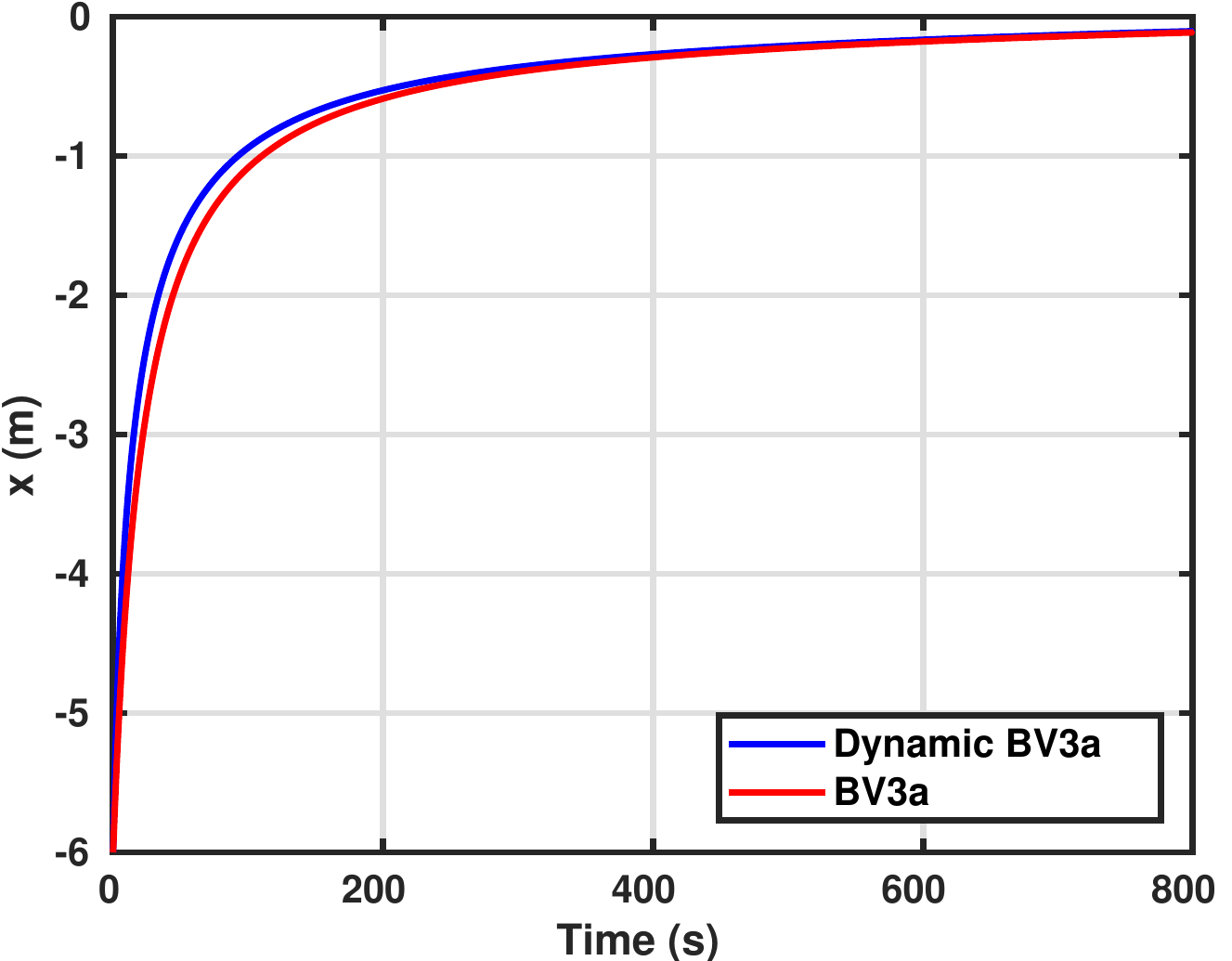}
    \caption{Trajectory over time of vehicles 3a and the proposed
      dynamic model with starting poses $x=-6$, $y=0$ and
      $\theta=0$} \label{fig:2}
  \end{center}
\end{figure}

To illustrate the increase of the speed when the vehicle points
towards the source -- prediction from equation (\ref{eq:Model}) -- we
simulated from the same initial position the standard Braitenberg
vehicle 3a and our new model version,
eq. (\ref{eq:bv3aDyn}). Specifically, the initial pose in both
simulations was $x=-6$, $y=0$ and $\theta=0$, which leads to a
straight line trajectory ($\dot{y}=0$ and $\dot{\theta}=0$) for our
stimulus.  Figure~\ref{fig:2} shows the time evolution of the $x$
coordinate, which corresponds to the solution of equation
(\ref{eq:dotX}). As we can see from the figure the version including
the dynamics (with $\alpha=3$) approaches the origin faster than the
standard Braitenberg vehicle 3a. We also concluded from the analysis
of equation (\ref{eq:Model}) that the forward speed in the dynamic
version is reduced when the vehicle points away from the source. This
effect will be shown in the next section together with the local
stability of the trajectories around the straight line trajectory.

\subsection{Stability close to the analytical trajectory}

\begin{figure}[h!]
  \begin{center}
    \includegraphics[width=0.97\columnwidth]{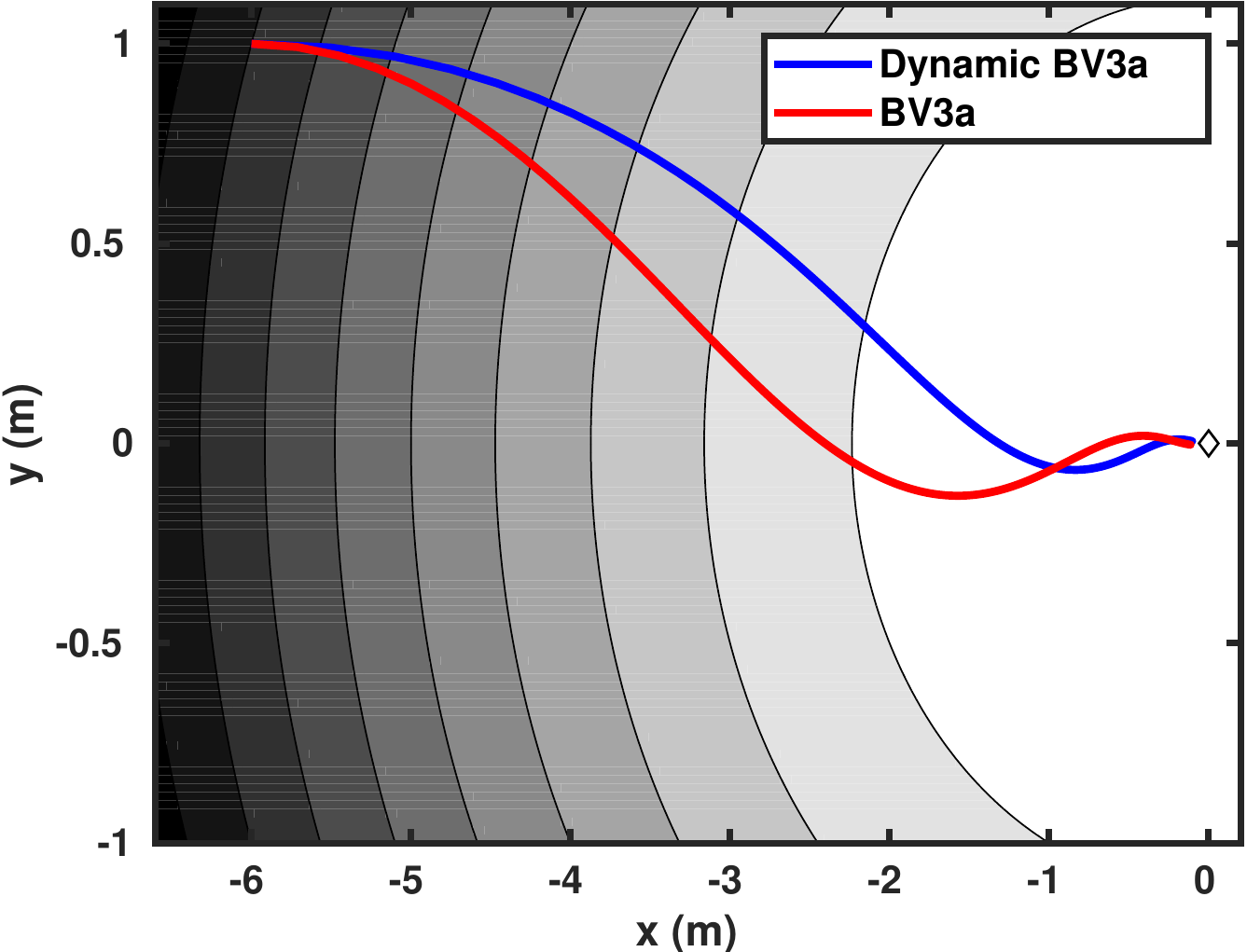}
    \caption{Simulation of the standard Braitenberg vehicle 3a with
      and without dynamic component with starting poses $x=-6$, $y=1$
      and $\theta=0$} \label{fig:3a}
  \end{center}
\end{figure}

As we saw in section~\ref{sec:parab} the real part of the eigenvalues
of the linearised system can get more negative for the Braitenberg
vehicle with the stimulus derivative controller.  That means the
convergence of trajectories towards the straight line solution is
faster. To illustrate this effect we simulated the Braitenberg vehicle
3a with and without the dynamic contribution for a starting position
close to the analytic straight line solution. Figure~\ref{fig:3a}
shows the resulting trajectories with starting pose $x=-6$, $y=1$ and
$\theta=0$. Since the stimulus is not circularly symmetric,
oscillations appear (see \cite{rano14results}), i.e. two eigenvalues
are complex conjugates, but the amplitude of the oscillations is
smaller for the dynamic version due to the exponential term (real part
of the eigenvalues) multiplying the oscillatory functions. The plot
also shows that the trajectory of the dynamic version of the vehicle
is shorter and seems to oscillate less around the linearisation
trajectory, which seems to indicate that the imaginary part of the
eigenvalues is also smaller.

\begin{figure}[h!]
  \begin{center}
    \includegraphics[width=0.97\columnwidth]{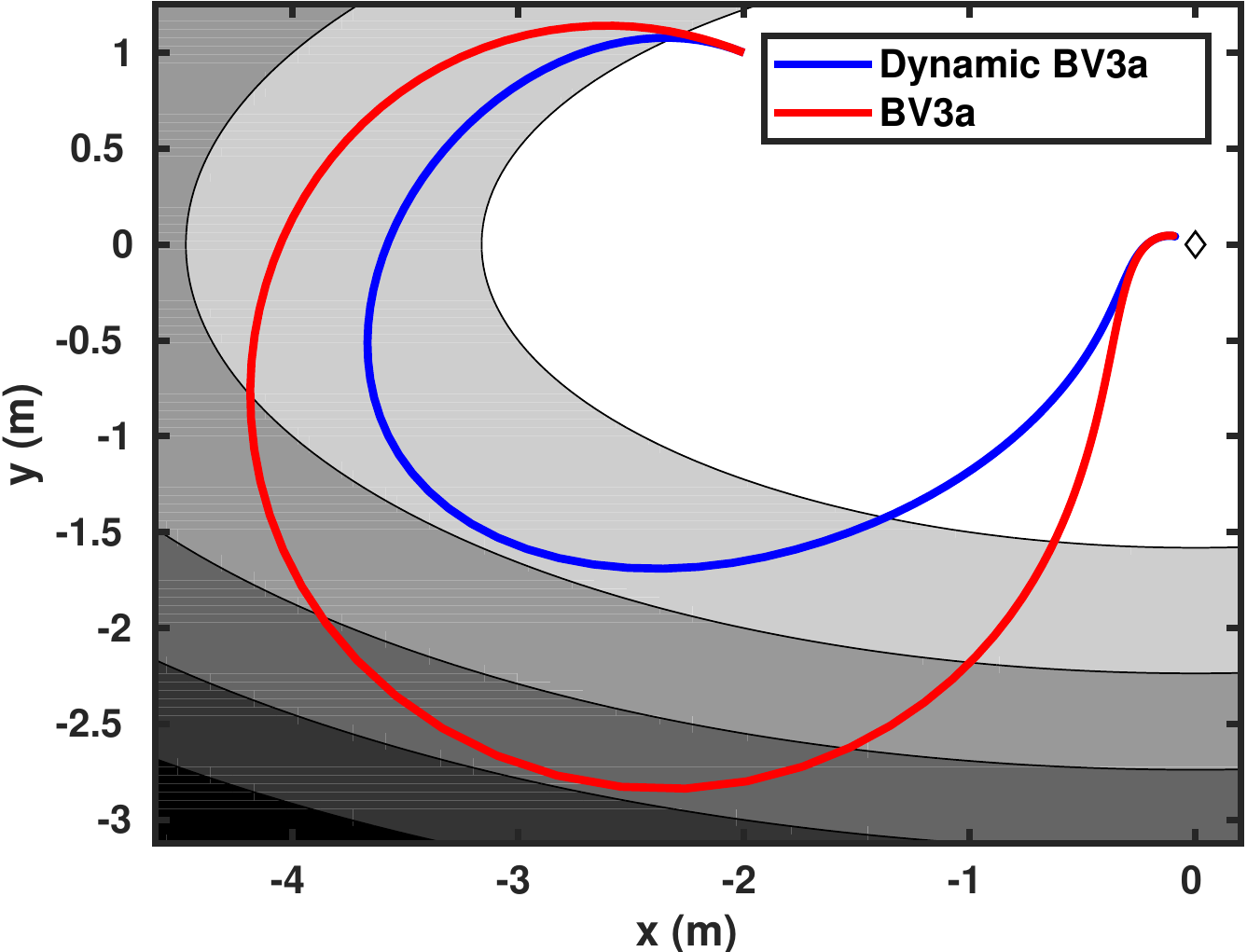}
    \caption{Simulation of the standard Braitenberg vehicle 3a with
      and without dynamic component with starting poses $x=-2$, $y=1$
      and $\theta=\arctan-\frac{1}{2}$} \label{fig:3b}
  \end{center}
\end{figure}

Finally, we ran a simulation to illustrate the effect on the reduction
of the forward speed when the vehicle points in the opposite direction
to the source, i.e. when $\nabla S^T\hat{e}<0$. According to equations
(\ref{eq:Model}) the speed should decrease compared to the standard
vehicle 3a while the vehicle moves opposite to the source. Once it
heads the direction of the gradient within $\pm\frac{\pi}{2}$ radians
the forward speed increases. Figure~\ref{fig:3b} illustrates this
effect by showing the trajectories of the vehicles starting with pose
$x=-2$, $y=1$ and $\theta=\arctan-\frac{1}{2}$, i.e. pointing directly
away from the source. As we can see the movement of the vehicle
greatly improves when the controller includes the time derivative of
the stimulus leading to a sharper turn and a shorter
trajectory. Although in this case the initial pose of the vehicle is
not close to the straight line trajectory, the vehicle firts turns
towards the source and the approaches the horizontal axis where the
equations of the linearised system are valid.
  
\section{Conclusions and Future Work}
\label{sec:conclusions}
This paper proposes a new bio-inspired local navigation strategy that
extends the well known Braitenberg models of animal positive taxes,
while accounting for the experimental findings in biology showing that
animals use the time evolution of the perceived stimulus to control
their movement. We analysed the mathematical model of the dynamic
Braitenberg vehicle and showed that the convergence towards the
stimulus source is faster. This work has important implications in
robotics as it shows the improvement control techniques based on
optical flow or event based cameras -- when the perception is
dependent on the speed of the robot -- can bring. Although the
circular dependency between the controlled variables and the perceived
variables makes the closed-loop system difficult to analyse, our
simplified setup enables applying analytical tools to show that there
is a clear improvement in the stability of the system.

One underlying assumptions to make the closed-loop system tractable is
the absence of sensor noise or a high signal to noise ratio. However,
biological systems are inherently noisy, and the time derivative of a
perceptual signal with additive noise cannot be used directly to
control a robot motion. It would require instead a band-pass filter.
Therefore, our next step will focus on trying to analyse the behaviour
of this new controller using a stochastic model of Braitenberg
vehicles \cite{rano17drift}.

\bibliographystyle{IEEEtran}
\bibliography{pequeno19sensors}

\end{document}